\documentclass[runningheads]{llncs}
\usepackage[T1]{fontenc}
\usepackage{amssymb}
\usepackage{amsmath}
\usepackage{graphicx}
\usepackage{booktabs} 
\usepackage{mathtools}
\begin{document}
\title{XpertAI: uncovering regression model strategies for sub-manifolds}
%
%
\author{Simon Letzgus\inst{1}\orcidID{0000-0003-0044-8959} \and
Klaus-Robert Müller \inst{1,2,3,4}\orcidID{0000-0002-3861-7685} \and
Grégoire Montavon\inst{2,5}\orcidID{0000-0001-7243-6186}}
\authorrunning{S. Letzgus et al.}
%
\institute{Machine Learning Group, Technische Universität Berlin, Germany \and
BIFOLD -- Berlin Institute for the Foundations of Learning and Data, Berlin, Germany \and
Department of Artificial Intelligence, Korea University, Seoul, Korea \and
Max Planck Institute for Informatics, Saarbrücken, Germany \and
Charit\'e -- Universit\"atsmedizin Berlin, Germany
}
\maketitle              
\begin{abstract}
In recent years, Explainable AI (XAI) methods have facilitated profound validation and knowledge extraction from ML models. While extensively studied for classification, few XAI solutions have addressed the challenges specific to regression models. In regression, explanations need to be precisely formulated to address specific user queries (e.g.\ distinguishing between `\textit{why is the output above 0?}' and `\textit{why is the output above 50?}'). They should furthermore reflect the model's behaviour on the relevant data sub-manifold. In this paper, we introduce \textit{XpertAI}, a framework that disentangles the prediction strategy into multiple output range-specific sub-strategies and allows the formulation of precise queries about the model as a linear combination of those sub-strategies. \textit{XpertAI} is formulated generally to work alongside popular XAI attribution techniques, based on occlusion, gradient integration, or reverse propagation. Qualitative and quantitative results demonstrate the benefits of our approach.
\end{abstract}

\section{Introduction}
Machine learning has provided powerful predictive models for numerous scientific and industrial applications. As the use of ML models for critical autonomous decisions increases, there is a growing demand for establishing trust while maintaining their predictive capabilities. Explainable artificial intelligence (XAI) has emerged as a step towards enhancing transparency and allows for insights into the inner workings of these highly complex AI models \cite{DBLP:journals/inffus/ArrietaRSBTBGGM20,DBLP:journals/pieee/SamekMLAM21}. XAI can be utilized for both, model validation against expert intuition as well as for obtaining new insights into the data-generating processes under investigation \cite{krenn2022scientific,klauschen2024toward}.

\begin{figure*}[h]
\centering
\includegraphics[width=0.9\linewidth]{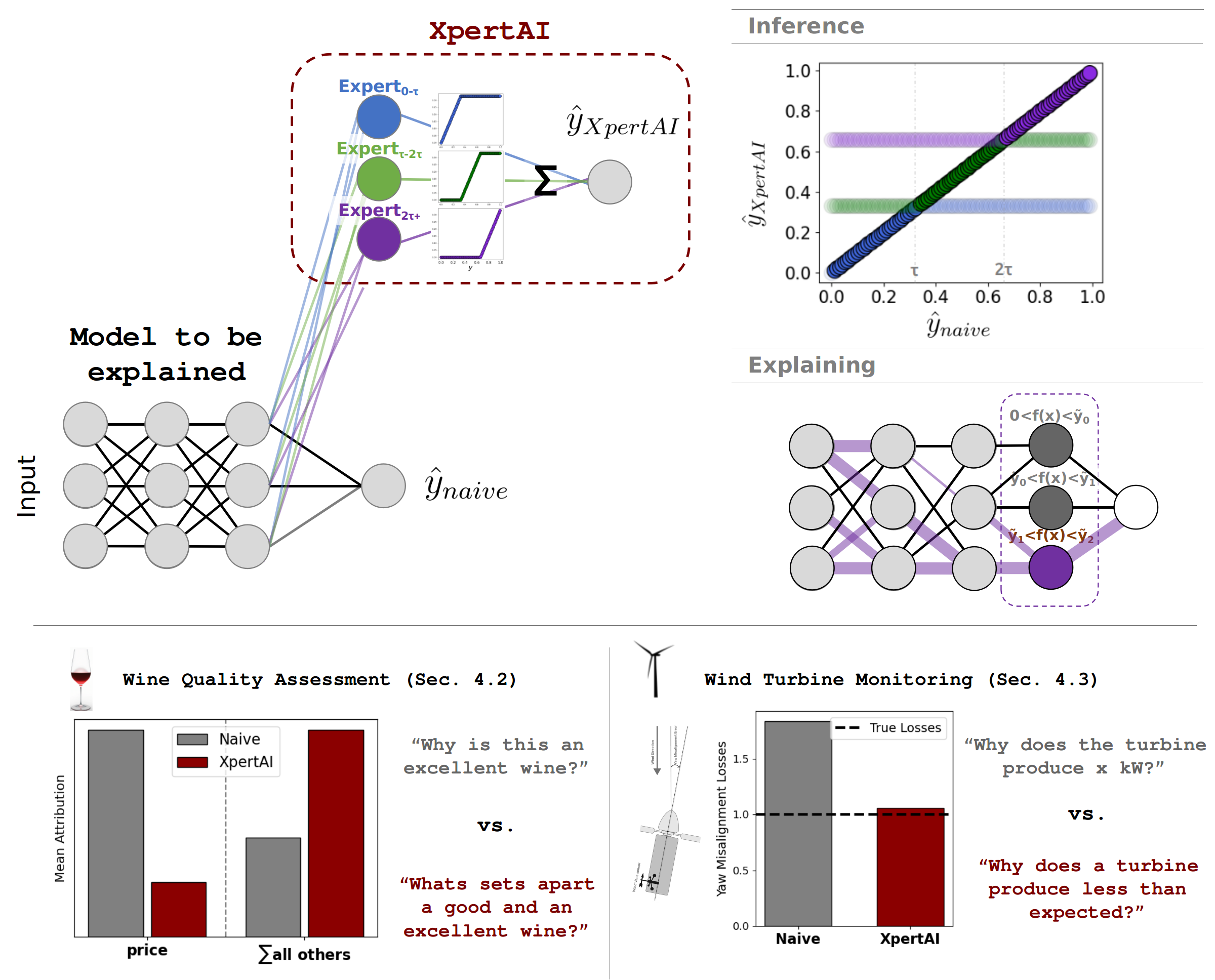}

    \caption{\textbf{Top:} Conceptual overview of our proposed \textit{XpertAI} approach. We add a layer of
\textit{range expert} neurons, each responsible for mimicking the original model behaviour on a range-specific sub-manifold of the data. During inference, the outputs of all range experts are added up and result in the original model output. When
explaining, we isolate output-range-specific effects by querying only the respective or a combination of range experts. \textbf{Bottom:} While the naive application of attribution methods typically answers questions from a generic point of view (grey) our approach enables answers to more nuanced questions as defined by the user (red). For the tasks of wine quality prediction and attributing losses of a wind turbine, we see significant structural changes in the explanations. For details see sections \ref{sec:wine_example} and \ref{sec:wind_example}.}
    \vspace{-7mm}
    \label{fig:intro_overview}
\end{figure*}

So far, the predominant focus within XAI has been placed on understanding the decisions made by classification models \cite{baehrens2010explain,ZeilerFergus,bach-plos15,Zhou16,DBLP:conf/kdd/Ribeiro0G16}. The widely used family of post-hoc attribution methods aims to achieve this by allocating evidence for a particular class across the corresponding input features. In doing so, they indicate the extent to which each feature has contributed to the model output. In this process, the model's decision boundary serves as a natural point of reference for the explanation. In regression, on the other hand, the equivalent to the decision boundary needs to be defined for every single query, since it is a priori unknown which of the two questions `\textit{why is the output above 50?}' or `\textit{why is the output above 0?}' is most relevant for the user \cite{Letzgus2022XAIRieee}. Moreover, in non-linear problems, sub-manifolds on which the model builds specific responses are to be expected, for example, for different output values.

To address these challenges, we propose our \textit{XpertAI} framework. The basic idea is to decompose the output of the regression model into a set of additive basis functions, the so-called \textit{range experts} (compare Figure \ref{fig:intro_overview}, top). Each range expert is dedicated to capturing the model behavior within a specific, output-range-dependent sub-manifold. Subsequently, the user can query the range experts with {\em any} state-of-the-art attribution method to obtain explanations that are contextualized to the individual explanatory needs. We demonstrate the benefits of our method on several (controlled and real-world) problems (see Figure \ref{fig:intro_overview}, bottom). We, for example, find that a model considered the price the most important input feature to distinguish an excellent wine from a bad one. But when explaining with respect to decent alternatives (close-by-reference values), other quality-related features become much more important. In another case study, we used attributions to monitor the performance of a wind turbine. There, we find that our contextualized explanations more faithfully capture the performance losses, which enables better maintenance decisions in practice. In addition to these qualitative insights, we report improved faithfulness through better contextualization with \textit{XpertAI}. An implementation is available online.\footnote{https://github.com/sltzgs/XpertAI}

\section{Related Work}
\label{section:relatedwork}

Our proposed method relates to several specific areas of XAI, which we will briefly discuss within this chapter (see e.g.~\cite{DBLP:journals/pieee/SamekMLAM21,DBLP:journals/inffus/ArrietaRSBTBGGM20} for XAI reviews). 

\subsection{Mixture of Experts}
The Mixture of Experts (MoE) framework \cite{JacobsMixExp91,142911,pawelzik1996annealed} follows a divide-and-conquer strategy, commonly used to enhance model performance. Recent work has applied MoEs for transparency by combining interpretable linear experts \cite{ismail2023interpretable}. In contrast, our approach utilizes MoEs for explaining models in a post-hoc manner, without restriction on the structure of the model, and steering the expert to become `range experts' focusing on specific value ranges. This is achieved by dividing the data into sub-manifolds according to the output range of a regression model, a way of domain-informed gating, and explaining the model strategy within these specific regions.

\subsection{Context in XAI attribution methods}
\label{sec:related_context}
Generally speaking, every explanation requires context to be meaningful. When explaining the outcome of a classification model, the decision boundary serves as a natural point of reference. Contrastive explanations have been proposed to better incorporate user-specific context into the explanation \cite{jacovi2021contrastive,stepin2021survey,lucic2020does}. For regression models, on the other hand, explanations depend on the reference output relative to which we seek an explanation \cite{Letzgus2022XAIRieee}. XAI attribution methods allow for the incorporation of context through baselines, which depending on the method have to be chosen in input space \cite{DBLP:conf/nips/LundbergL17,DBLP:conf/icml/SundararajanTY17} or latent space \cite{DBLP:conf/icml/ShrikumarGK17,DBLP:journals/pr/MontavonLBSM17,Letzgus2022XAIRieee}. Each baseline then corresponds to a respective reference value ($\widetilde{y}$). In practice, the choice of baselines represents a challenge with fundamental impact on the outcome of the explanation. In this work, we therefore propose a practical solution that ensures contextualization by design for regression models and, as a result, increases robustness against suboptimal baseline choices. 

\subsection{Disentangled XAI and Virtual Layers}
While refining the question to be asked is essential in a regression setting, many works have focused on independently refining the explanation itself (mainly in a classification context). Specifically, enriching explanations by identifying its multiple components, associated with distinct abstract concepts. These can be obtained in a supervised manner \cite{kim2018interpretability,zhao2023counterfactual}, in an unsupervised manner \cite{vielhaben2023multidimensional,chormai2022disentangled}, or by directly inspecting neurons \cite{ZeilerFergus,Zhou16,achtibat2022where}. This kind of analysis often involves an informed transformation of latent representations to obtain a meaningful or relevant 'concept space', followed by the inverse transformation to leave the overall model behaviour intact \cite{wang2024disentangled}. Therefore, these approaches are referred to as \textit{virtual layers}. \cite{chormai2022disentangled}, for example, extract sub-concepts that jointly contribute to the explanation of an overall class concept. Likewise, \cite{vielhaben-fourier} generates a Fourier basis on which the prediction of speech samples can be analyzed more efficiently, and \cite{linhardt2024preemptively} introduces a virtual PCA layer, which disentangles verified from unverified factors of variation and subsequently prune the latter for increased robustness. We extend these efforts to the broad domain of regression, by introducing a novel technique that aims to disentangle global phenomena that exert influence consistently across the entire range of potential regression outputs from more localized context-specific patterns (see Figure \ref{fig:intro_overview}, top).

\section{Our Method: XpertAI}
\label{sec:method}

\begin{figure*}[h]
\centering
\includegraphics[width=0.95\textwidth]{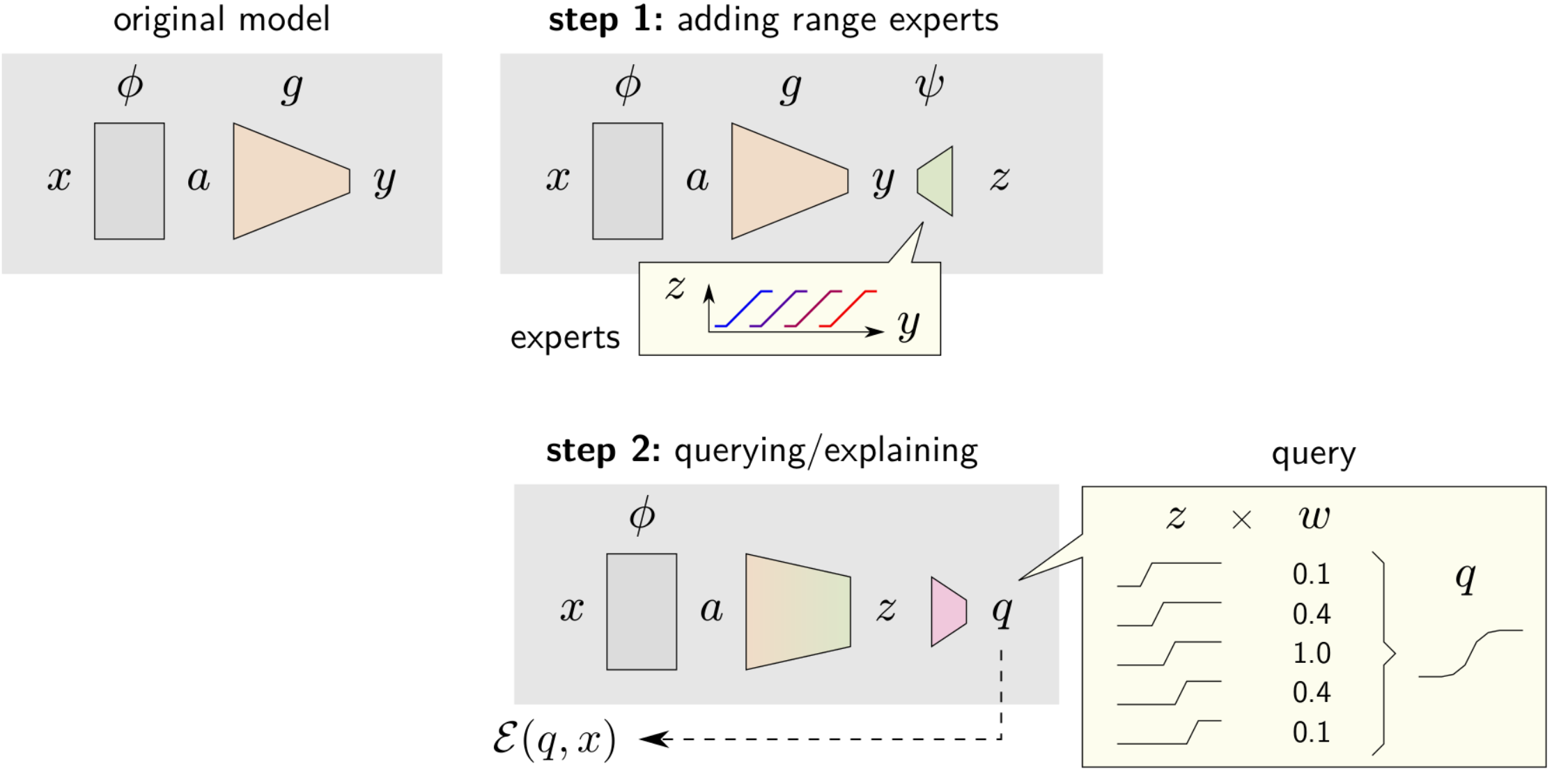}
\caption{Diagram of our two-step approach for obtaining fine-grained explanations from an existing regression model. The first step consists of adding a collection of range experts to the model. The second step synthesizes a query $q$ from those range experts and produces a corresponding explanation (the exemplary query on the right is sigmoidal with the ML model's output but linear with the experts).}
\label{fig:models}
\end{figure*}

In the following, we introduce our novel method, called \textit{XpertAI} for explaining neural network regression models. Our approach is inspired by the MoE concept and consists of appending \textit{range experts} to a given ML model, thus allowing the user to formulate precise \textit{queries} for which range they need an explanation. This appendage can be seen as a virtual layer inserted in the neural network, which -- while leaving the overall prediction function intact -- enriches it by providing the basis for query formulation and explanation. Figure \ref{fig:models} conceptually depicts the method and its notation, with details in the following sections.

\subsection{Adding Range Experts}
\label{section:rangeexperts}

We abstract the ML model as a function $f$ mapping the input $x$ to a real-valued output $y$. The model may either be a pure black-box or a neural network with multiple layers. We define the range experts as the following collection of functions building on the output of the ML model:
\begin{align}
z = 
\left(
\begin{array}{l}
\rho_{0,\tau}(y)\\
\rho_{0,\tau}(y-\tau)\\
\rho_{0,\tau}(y-2\tau)\\
\vdots
\end{array}\right)
\label{eq:experts}
\end{align}
where $\rho_{0,\tau}(y) = \min(\max(y,0),\tau)$ clips the input to the interval $[0,\tau]$. A low $\tau$ corresponds to more specialized experts. The kind of transformation in Eq.\ \eqref{eq:experts} is also known as thermometer coding. The architecture that results from appending these experts is shown in Fig.\ \ref{fig:models}. Assuming the values of $y$ are always positive (which we can ensure through offsetting) we can reconstitute the output prediction by summing the experts' outputs:
\begin{align}
\textstyle y = \sum_m z_m
\label{eq:identity}
\end{align}
The mapping from $y$ to $z$ and back to $y$ can be seen as a virtual layer which does not affect the input-output mapping, but that provides additional functionality. Unlike previous formulations of virtual layers \cite{vielhaben-fourier}, ours is placed at the output, enabling a disentanglement of the explanation in terms of output ranges.

Consider now the task of attribution. Classical explanation techniques would attribute $y$ to the features of $x$ (something we denote by $\mathcal{E}(y,x)$). The virtual layer allows us to compose two attribution steps:
\begin{align*}
R_m &= \mathcal{E}(y,z)_m\\
R_{im} &= \mathcal{E}(R_m,x)_i
\end{align*}
where $R_m$ denotes the contribution of expert $m$ to the output $y$ (in our case we simply have $R_m = z_m$), and $R_{im}$ can be interpreted as the contribution of input feature $i$ through expert $m$. The overall explanation can be seen as a matrix of size \# features $\cdot$ \# experts, from which it will be possible to formulate and answer precise user queries.
\subsection{Querying\,/\,Explaining}
\label{section:query}
Whereas the disentanglement performed above provides a more detailed view of the prediction behaviour than a simple explanation, the user is often interested in particular aspects of it. Our approach lets the user formulate a query (or `explanandum') as a linear combination of the range experts:
\begin{align}
\textstyle 
q = \sum_m w_m z_m
\label{eq:query}
\end{align}
An example of such a query is given in Fig.\ \ref{fig:models} (right).  For example, if the user is interested in what makes a prediction $y = 60$ larger than a reference value of $50$, the query $q$ can be shaped in the form of a sigmoid centred at the reference value $50$.

Once a query has been prepared (i.e.\ once the weights $w_m$ have been defined), an explanation to that query $\mathcal{E}(q,x)$ can be generated by any state-of-the-art attribution method:
\begin{align}
\textstyle
\mathcal{E}(q,x) = \mathcal{E}\big(\sum_m w_m z_m,x\big)
\label{eq:explanation}
\end{align}
Note that for explanation techniques that fulfil the linearity axiom w.r.t.\ the last layer of representation, we can further develop the expression of the explanation as:
\begin{align}
\mathcal{E}(q,x) = \sum_m w_m \mathcal{E}\big(z_m,x\big)
\label{eq:explanation-sum}
\end{align}
It shows that the explanation is a linear combination of the explanations of all basis elements $z_m$. This formulation can be advantageous when the explanation is associated with many different queries or when the query arrives in real-time, in which case the explanation basis can be pre-computed. We note that our approach satisfies some key desirable properties of an explanation:

\begin{proposition}[Conservation]
\label{proposition:conservation}
If $\forall_m: \sum_i \mathcal{E}(z_m,x)_i = z_m$, then $\sum_i \mathcal{E}(q,x)_i = q$, in other words, if each range expert $z_m$ can be attributed to input features in a conservative manner, then explanations of any query $q$ are also conservative.
\end{proposition}

\begin{proposition}[Irrelevance]
\label{proposition:irrelevance}
If $\forall_m: \mathcal{E}(z_m,x)_i = 0$, then $\mathcal{E}(q,x)_i = 0$, in other words, if we verify that for a given data point, the feature is irrelevant for all range experts, then it is also irrelevant for any query built on those experts.
\end{proposition}

These two results are easily retrievable by observing the specific structure of the explanation given in Eq.\ \eqref{eq:explanation-sum}. Proofs can be found in Appendix \ref{app:proof_proposition}.

\subsection{Structural Disentanglement}
\label{section:technical}

When the underlying explanation method relies not directly on the ML model's output but on its computational graph (e.g.\ LRP), the latter must be disentangled. Clearly, such a structural disentanglement is missing as the mapping from activations $a$ to the expert's outputs $z$ passes through a one-dimensional bottleneck $y$ (the original real-valued output). We propose to replace the original mapping $a \mapsto (z_m)_m$ by a learned surrogate model $(s_m)$:
$$
a \xmapsto{\theta} (s_m)_m \mapsto (\widehat{z}_m)_m
$$
where the second part of the mapping is given by $\widehat{z}_m = \rho_{0,\tau}(s_m)$, a hard-coded saturation forcing the surrogate and true experts to produce outputs in the same range. We then build for each expert the loss function:
$$
\ell(s_m,z_m) = 
\left\{\begin{array}{ll}
\max(0,s_m) & z_m \leq 0\\[.5mm]
|s_m - z_m| & 0 < z_m < \tau\\[.5mm]
\max(0,\tau-s_m) & z_m \geq \tau
\end{array}\right.
$$
which encourages that the surrogate's output is correct within-range and on the correct side outside-range. We then solve $\min_{\theta} \mathbb{E}[\sum_{m} \ell(s_m, z_m)]$ with $\mathbb{E}[\cdot]$ denoting the expectation over the training data. To preserve not only the prediction output of the original model but also its prediction strategy (i.e.\ the feature it uses) further steps are needed. One approach is to enforce the loss function not only on the data but also on perturbations of the data \cite{Stutz_2019_CVPR}. For example, activations can be randomly turned off (with a probability chosen between $0$ and $1$). This perturbation scheme ensures in particular that the Shapley value explanations of the original and disentangled models become similar (i.e.\ that they predict the same for the same reasons). Furthermore, we find that freezing the bias in the output layer is important to achieve the desired structural disentanglement.

\subsection{XpertAI evaluation}
\label{section:evaluation}
We evaluate our proposed approach qualitatively (Section \ref{sec:qual_eval}) and quantitatively (Section \ref{sec:quant_eval}). In both cases we rely on either a (constructed) problem that allows for validation against some sort of ground truth, or the observation of model behaviour under attribution-guided, meaningful input perturbations. \cite{hama2022deletion} proposed a regression-specific metric called the area between the curves (ABC). The ABC is defined as the area between the model output when occluding a sample's features in the order of attribution magnitudes and a straight line connecting $f(x)$ and $f(x')$ (which corresponds to random sorting). Since sorting ascending and descending can result in asymmetrical curves, we sum over both areas \cite{bluecher2024decoupling}. For a balanced result, we normalize by the distance between the sample and the baseline when averaging. Higher values of ABC are better. 

Furthermore, the challenge of including context in attribution methods (Sec. \ref{sec:related_context}) naturally extends to occlusion-based evaluation (ergo, what to occlude with?). To ensure that we evaluate attributions within the relevant output range of function $f(x)$, where we account for context-specific (local) effects, we occlude with a domain-specific counterfactual \cite{Albini_2022,hama2022deletion}. Therefore, we sample conditional $x'=D(x|y=\widetilde{y})$ from the available data set $D$ with which we then occlude and average the respective ABCs over multiple draws.

\section{XpertAI-Opinion: insights into model behaviour on sub-manifolds}
\label{sec:qual_eval}
We now demonstrate how our \textit{XpertAI} approach can help users disentangle local and global effects for meaningful insights in different case studies. First, we uncover output-scale-specific strategies for image regression problems (\ref{sec:mnist_example}). Then, we explain the quality of red wine (\ref{sec:wine_example}) and the production losses of a wind turbine due to a technical malfunction (\ref{sec:wind_example}). For each of the problems, we briefly introduce the dataset, model and \textit{XpertAI} setting, before presenting the insights. We present results from using both, Integrated Gradients and Layer-wise Relevance Propagation (LRP). Details on all case studies can be found in Appendix \ref{app:details_case}.

\begin{figure}[h]
\centering
\includegraphics[width=0.8\linewidth]{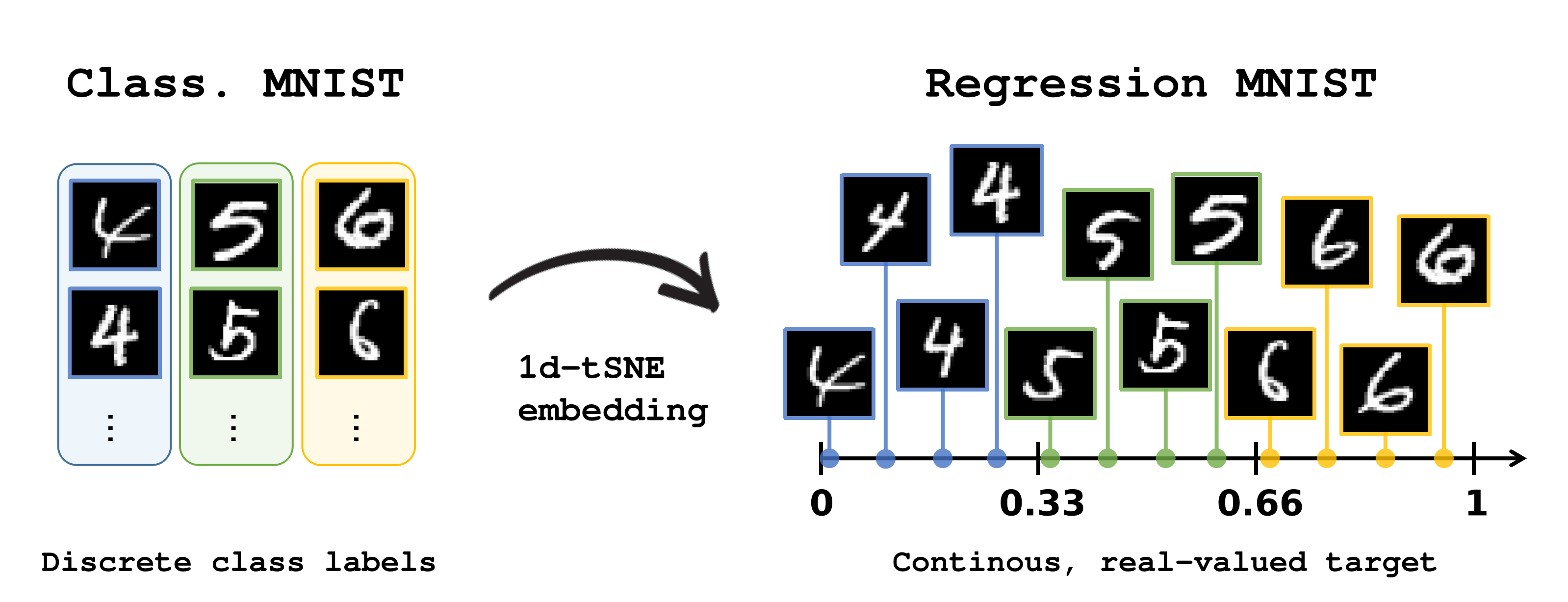}
    \caption{Examples from three classes of the MNIST dataset for handwritten digit recognition (left) mapped to a real-valued scale with the help of a one-dimensional t-SNE embedding (right). Digits populate continuous ranges of the new target, and sorting within the digit ranges corresponds to digit rotations.}
    \label{fig:dataset_mnist}
\end{figure}

\subsection{Uncovering output-scale-specific strategies}
\label{sec:mnist_example}

First, we adopt the well-known MNIST \cite{deng2012mnist} dataset and transform it into a regression problem (\textit{rMNIST}). For simplicity, we take the subset of only three digits (4,5, and 6) and calculate a one-dimensional t-SNE representation \cite{van2008visualizing}, which henceforth serves as a new label for each sample. Additionally, we ensure labels are distributed uniformly between values of zero and one. As a result, the individual digits populate continuous parts of the output dimension (in our case sorted by digit magnitude, which facilitates interpretation) while sorting within each digit bin is based on the respective digit's rotation (compare Figure \ref{fig:dataset_mnist}). We now train a vanilla CNN model architecture to learn this mapping from image to output scale. For contextualized insights, we train three range experts (one for each digit range). We first discuss qualitative results and present its quantitative evaluation in Section \ref{sec:quant_eval}.

\begin{figure}[h]
\centering
\includegraphics[width=\linewidth]{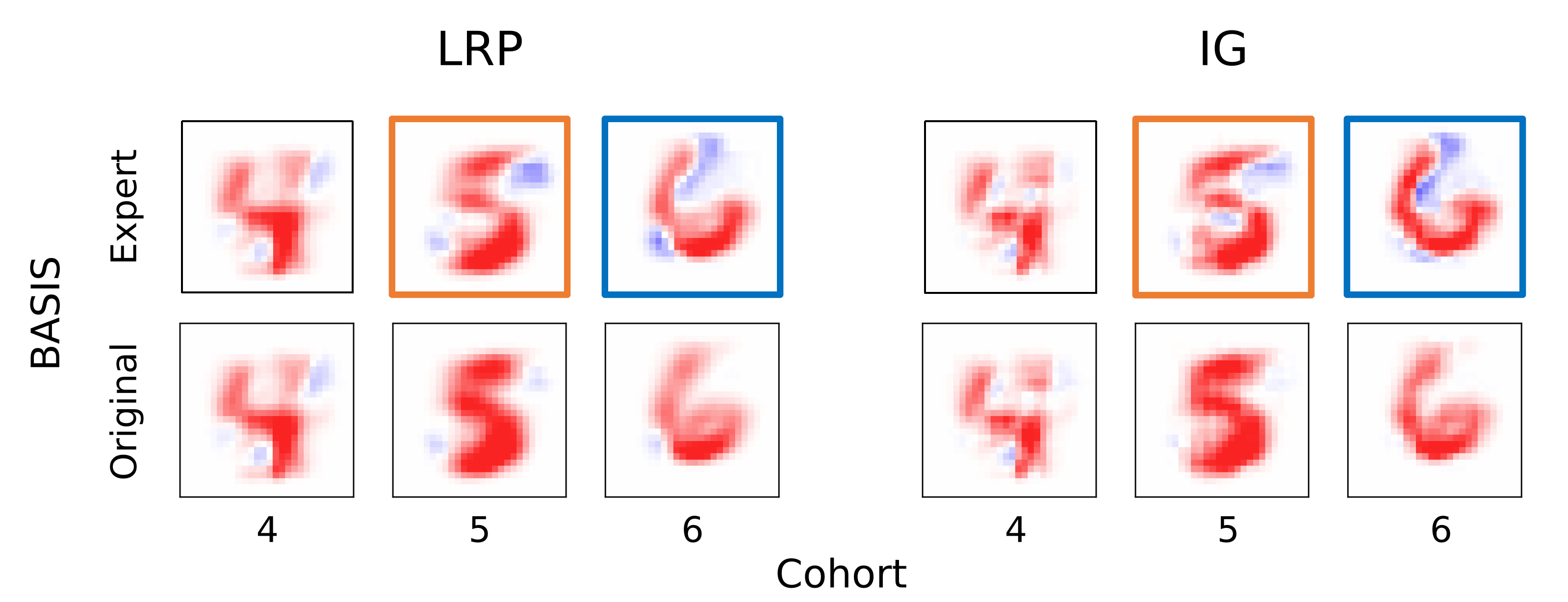}
    \caption{Mean attributions over different cohorts of samples (columns) and basis functions (rows). The bottom row represents naive attributions. The top row corresponds to the respective range-specific expert \textit{XpertAI} bases. Note, how only the latter exposes the digit rotation within the digit ranges (orange/blue).}
    \label{fig:mean_att_mnist_three}
\end{figure}

Figure \ref{fig:mean_att_mnist_three} shows a comparison between the standard and the \textit{XpertAI} explanations for both, LRP and IG. We contrast the average naive attributions over all samples \textit{within} the respective output range (bottom row) with the explanations obtained with the respective range experts (top row). The explanations for the digit range \textit{4} remain the same since both implicitly assume the same reference value (zero on the output scale). The expert attributions for the upper digit ranges (marked in orange and blue), enable more granular insights. It is visible how the range experts focus specifically on the rotation of the digit: a rotation to the right is associated with lower values (negative attribution, blue) and vice versa. See Appendix \ref{app:details_case} for more basis functions. In Section \ref{sec:quant_eval}, we will see that these qualitative differences in attributions also result in improved quantitative evaluation scores for attribution faithfulness.

\begin{figure}[h]
    \centering
    \includegraphics[width=0.65\linewidth]{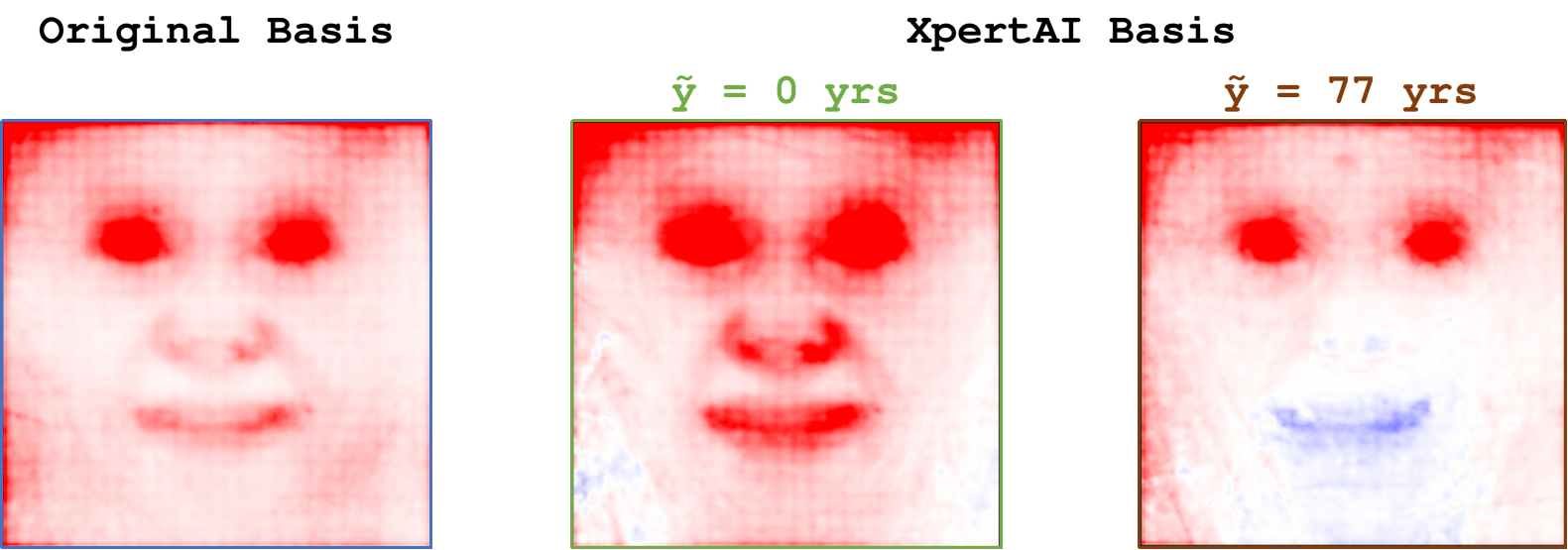}
    \caption{Comparison of average attributions for standard LRP (left) and two different XpertAI basis functions. Red indicates positive, and blue negative evidence. We can see that the disentangled explanations allow for much more fine-grained conclusions. Interestingly, the sign flip of the mouth area was masked by the strong attributions with respect to the original basis. We test for its faithfulness in Figure \ref{fig:face_masked_effect}.}
    \label{fig:face_explanations}
\end{figure}

Now, let's consider an illustrative regression task closer to real-world applications: biological age estimation from facial images \cite{han2013age,angulu2018age,abdolrashidi2020age} (see Appendix \ref{app:details_case} for model and data set details). Intuitively, the explanation for a person with a high age should be structurally different when being contrasted with a much younger age or an only slightly younger one. We, therefore, focus on a high-age cohort (individuals predicted to be above 77 years) and train three range experts ($\tau$ = 38.5 years). Figure \ref{fig:face_explanations}, left, shows LRP attributions for the original model, averaged over the respective samples. Our proposed approach now allows us to disentangle these further using the respective age-specific basis functions. As expected, the explanation relative to the 'young' basis (centre) overall contains much more positive evidence than the one with respect to the closer reference value of 77 years (right). Additionally, we can see that the latter is more fine-grained with a remaining focus on the person's eyes and, surprisingly, we discovered a sign-flip for the oronasal region with a particular focus on lips and teeth. While the mouth has been reported to be an area particularly vulnerable to biases in age estimation from facial images \cite{ganel2022biases}, we put the faithfulness of these particular explanations to the test. 

\begin{figure}[h!]
    \centering
    \includegraphics[width=0.75\linewidth]{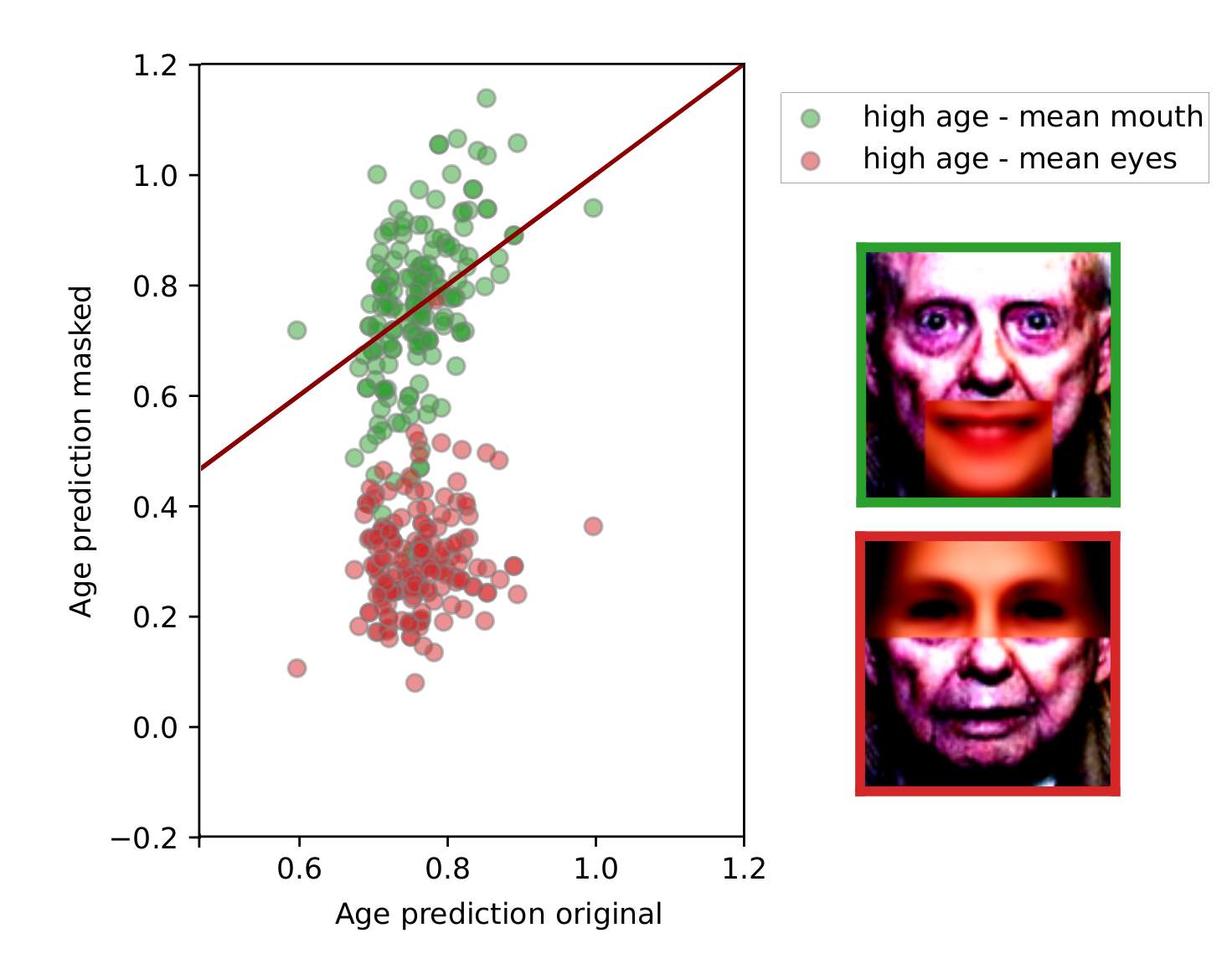}
    \caption{Validation of findings from disentangling age prediction (compare Fig. \ref{fig:face_explanations}). We occlude relevant parts of the image according to the disentangled explanations (eyes and mouth) with the dataset-wide average face (two examples on the right). We then observe the effect on the model output relative to the original model prediction. For the high-age cohort, the two areas have distinctly different effects. Occluding the eyes with a relatively younger pair results in a consistent decrease in the predicted age. Occluding the mouth region, however, results in an increase for many of the samples. This model behaviour is in line with our insights from the disentangled explanations.}
    \label{fig:face_masked_effect}
\end{figure}

In Figure \ref{fig:face_masked_effect}, we compare the effect of occluding the respective parts (eyes and mouth) of people's faces with a generic average over all images. One example of each is shown at the right of the figure. Recall that this means we mask the eyes and mouth section with a relatively 'younger' version. The chart shows the respective change in the model's output. In line with intuition, and the explanations, age is indeed consistently decreased when occluding the eyes. Masking the mouth area with relatively 'younger' mouths, however, indeed results in an \textit{increase} of the model's average prediction in many cases. The \textit{XpertAI}-basis therefore constitutes the more faithful explanation since the attributions correctly captured the sign flip in model behaviour.

In conclusion, the disentangled basis explanations enabled more detailed insights into the model's inner workings for both, the rMNIST and the age-prediction cases. They revealed effects that were not apparent from the naive explanations of the original model, since their highly aggregated nature did not allow for more fine-grained insights.

\subsection{What sets apart a good wine from an excellent one?}
\label{sec:wine_example}
As noted in the introduction, we now explore a more hedonistic and tangible example - red wine quality. We utilize Kaggle's \textit{Spanish red wine dataset}\footnote{https://www.kaggle.com/datasets/fedesoriano/spanish-wine-quality-dataset} which contains several thousand wine samples. They are described by five numerical (year, price, as well as body, acidity, and quality scores) and four categorical (name of the winery and the wine, grape, region) features. The quality score, which is an 'average rating' given by thousands of testers (rating binned into 8 discrete quality levels), is our regression target. After data-pre-processing around 1700 samples are left. We have trained a small fully-connected ANN which achieved an $R^2$ of around 0.7.

\begin{figure}[h]
\centering
\includegraphics[width=0.8\linewidth]{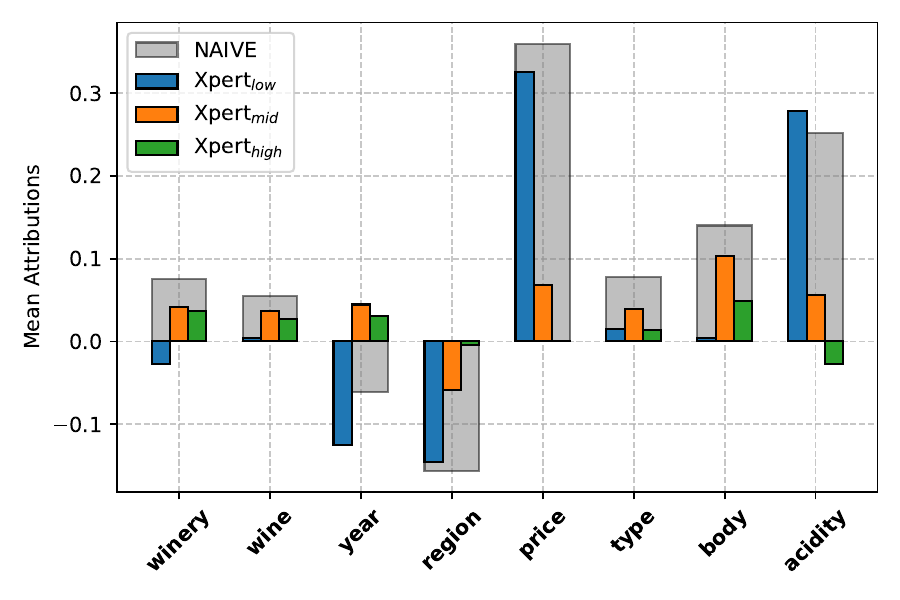}
    \caption{Decomposition of naive explanations (grey) for samples from the high output range ('excellent' wines) with respect to low, medium and high-quality reference values (colourful). The XpertAI explanations allow for nuanced insights into what makes an excellent wine better than the worst (blue), a decent (orange) or a good (green) alternative.}
    \label{fig:wine_bases}
\end{figure}

We now want to learn what, according to the model, sets apart a good wine from an excellent one. We define wine as good when it belongs to the top 10 \% and excellent when it belongs to the top 1.5 \% of the model output range. We train three range experts ($\tau=0.33$) and compare the respective attributions obtained from standard IG with its application within the XpertAI framework. Figure \ref{fig:wine_bases} shows the decomposition of the excellent wine attributions into the respective expert bases. Aside from the natural change in attribution scales, the most prominent difference in the explanations is the contribution of the price to the model outcome. For the naive IG attributions, the price is the by far most important feature (meaning high prices alone are the main indicator for excellent wines). The contextualized \textit{XpertAI} attributions, on the other hand, give a much more balanced picture. Here, the outcome suggests that the price is the most important feature only for the low-quality range expert (meaning what distinguishes an average from a poor wine, blue). The relative importance of the price, however, is significantly reduced when compared to average wines (orange) and almost vanishes when compared to good wines (green). There, the sum of all other quality criteria is much more important than the price of the wine itself. This directly translates to some actionable (and intuitive) insight: if you next time buy a wine in the supermarket, don't go cheap to ensure you buy a decent wine. When looking for an excellent one though, you might be better off with the expert judgement of your local wine seller.

To make sure, our insights are not based on intuitive but unfaithful attributions, we also compare quantitative faithfulness for the and observe an average increase in the ABC metric by more than 10 \% (see Sec. \ref{sec:quant_eval}).

\subsection{Why does the wind turbine produce less than expected?}
\label{sec:wind_example}
Wind power is one of the pillars of decarbonizing energy systems around the world. Wind turbines are often placed in remote locations and need to be operated and monitored from a distance, using data from their Supervisory Control and Data Acquisition (SCADA) system. Effectively leveraging this data is an active area of research \cite{doi:10.1080/14786451.2021.1890736}, with the primary focus on detecting and diagnosing underperformance as the central challenge \cite{pandit2023scada}. However, the detection of \textit{under}performance is always context-specific since the implicit question is: `underperformance relative to what operational state?' In the wind turbine case, it is the condition without the presence of a malfunction, given the context of prevailing ambient conditions.

We utilize data from a 2 MW wind turbine and a meteorological met-mast from an onshore wind farm on the Iberian peninsula \footnote{https://opendata.edp.com}. SCADA data is available for two years and includes ambient conditions as well as technical turbine parameters as 10-minute averaged values (50,000 data points after pre-processing). We have trained a small fully-connected MLP to predict the turbine output from wind speed, air density, and turbulence intensity. The model achieves a competitive RMSE of less than 36 kW. Additionally, we have augmented the data with so-called yaw-misalignment losses. They occur when a turbine does not perfectly face the incoming wind direction, which reduces the effective area of the rotor. Detecting yaw-misalignment is an ongoing field of research \cite{pandit2021operational,QU2022115786} and attributing it by XAI methods has recently been proposed as an effective solution \cite{Letzgus24XAIwind}. For such an approach to work, we need our XAI methods to faithfully attribute the losses induced by yaw-misalignment to the respective feature (difference of nacelle and wind direction). In our setup, we can directly compare attributions with the respective ground-truth-losses.

\begin{figure}[h]
\centering
\includegraphics[width=0.7\linewidth]{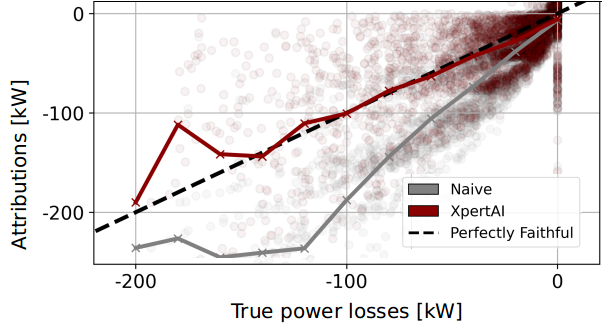}
    \caption{Quantitative faithfulness when attributing yaw-misalignment losses to the respective feature with standard LRP (grey) and \textit{XpertAI}-LRP (red) against the true losses (dashed line).}
    \label{fig:wind_results}
\end{figure}

We trained three range experts across the different operational regions of the turbine (see Appendix). Figure \ref{fig:wind_results} shows the comparison of attributing the yaw-misalignment induced losses to the respective yaw-feature with standard LRP and our XpertAI-LRP variant. We can observe that the naive LRP application attributes exhibit a systematic overestimation (larger negative values) of losses caused by incorporating phenomena from outside the respective operational regime. Our proposed novel attributions obtained from the range-experts, on the other hand, are on average much closer to the ground truth. For turbine operators, this directly translates to better operation and maintenance decisions and therefore highlights the benefit of using sub-manifold-specific explanations in industrial or engineering applications.

\section{Quantitative Evaluation and Sanity Checks of XpertAI-faithfulness}
\label{sec:quant_eval}
After having presented some intriguing insights enabled through our \textit{XpertAI} approach in the previous chapter, we now conduct a systematic evaluation of explanation faithfulness. Details on the respective experiments and additional insights for obtaining faithful range experts can be found in Appendix \ref{app:best_practice} and \ref{app:details_case}.

\subsection{Are XpertAI attributions faithful?}
To answer this question quantitatively, we utilize the ABC score as introduced in Section \ref{section:evaluation}. Table \ref{tab:res_overall} reports the ABC scores of our \textit{XpertAI} approach relative to a naive application of LRP and IG on the previously introduced data sets as well as several popular regression benchmarks \cite{scikit-learn}. For each of them, we trained three range experts and evaluated samples from the top range (see Appendix \ref{app:details_case}). Overall, we see consistent improvements in ABC scores across all settings which means that our approach indeed can generate more faithful attributions with respect to a user-specific query. Note, that the advantage is significantly larger for LRP where our approach corresponds to a data-driven root-search strategy whereas naively, there is no such option. For IG we have already leveraged its inherent contextualization capability to some extent by utilizing the mean over all input samples as a starting point for the integration path. Our approach is still able to further refine the attributions towards a better contextualization. 

\begin{table}[h]
\caption{Comparison of faithfulness for different attribution methods applied naively and within the \textit{XpertAI} framework. Relative improvement of ABC over naive application. Standard deviation over 5 different retraining runs for LRP.}
\label{tab:res_overall}
\vskip 0.15in
\begin{center}
\begin{small}
\begin{sc}
\begin{tabular}{l|ccc}
\toprule
dataset & $LRP$ & $IG$ \\
\midrule
rMNIST      & +50.7 \% $^{\pm 3.5}$  & +7.2 \% \\
WINE        & +19.8 \%  $^{\pm 1.2}$ & +10.6 \%          \\
FRIEDMAN    & +12.6 \%  $^{\pm 0.4}$  & +1.9 \%            \\
CALIFORNIA  & +2.5 \%  $^{\pm 0.9}$   & +9.7 \%            \\
DIABETES   & +3.8 \%  $^{\pm 1.6}$  & +4.4 \%               \\
\bottomrule
\end{tabular}
\end{sc}
\end{small}
\end{center}
\vskip -0.1in
\end{table}

\subsection{How many range experts?}
One practically relevant question is, how many range experts to train, which includes the choice of their respective ranges ($\tau$). Conceptually, the method works best if every distinct sub-region of the output is covered by at least one \textit{range expert}. In practice, these can either be domain-informed and therefore known apriori, or inferred by analyzing activation patterns (from an activation vs. $f(x)$ scatter plot, for example). In the context of our rMNIST case, selecting one range expert for each digit range, therefore three range experts in total appears to be the most intuitive choice. Since in practice, we might not know where exactly these boundaries lay, we compare settings for three, five, six, and nine equally spread range experts. 

\begin{table}[t]
\caption{Results for pixel flipping experiments for regression MNIST. Results within ranges: sample-flipping baseline pairs are within one expert range. \textit{ABC} values are normalized by flipping distance. Values for naive methods differ because of the normalization. High values are better.}
\label{tab:res_MNIST_intra}
\vskip 0.15in
\begin{center}
\begin{small}
\begin{sc}
\begin{tabular}{c|cc|cccr}
\toprule
$\# $ &$LRP$ & $LRP$ & $IG$ & $IG$ \\
$experts $ & $Naive$ & $XpertAI$ & $Naive$ & $XpertAI$ \\
\midrule
 3 & 0.40   & \textbf{0.56}$\pm$ 0.02      & 0.93& \textbf{1.00} $\pm$ 0.05\\
 5 & 0.47   & \textbf{0.70}$\pm$ 0.01      & 1.16& \textbf{1.20} $\pm$ 0.01\\
 6  & 0.46    & \textbf{0.75}$\pm$ 0.01      & 1.22& \textbf{1.23}$\pm$ 0.01 \\
 9  & 0.49    & \textbf{0.78}$\pm$ 0.01      & 1.32& \textbf{1.36}$\pm$ 0.01 \\
\bottomrule
\end{tabular}
\end{sc}
\end{small}
\end{center}
\vskip -0.1in
\end{table}

In Table \ref{tab:res_MNIST_intra} we see that our approach improved the ABC score across all settings. Also, we can see that LRP benefits in particular from adding extra range experts while IG results are more consistent across the number of experts. Note, that this also holds if the expert ranges are not aligned with known sub-concept ranges (as is the case for 5 equally distributed experts). In practice this means that the limit for the number of experts depends on the specific problem, computational considerations as well as the resolution of the available data.

\subsection{(Diss-)aggregate XpertAI-attributions}
\begin{figure}[h]
\centering
\includegraphics[width=0.7\linewidth]{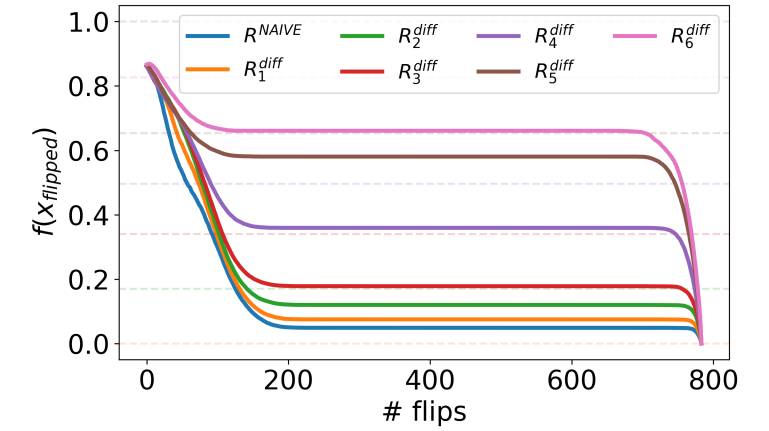}
\vspace{-5mm}
    \caption{Mean occlusion curves over all samples from the top bin of a six-expert-basis. When successively subtracting range-expert attributions from the original explanation and flipping pixels according to the remaining explanations, the flipping curves saturate in the proximity of the respective reference values.}
    \label{fig:flip_curves}
\end{figure}

From \textit{Proposition 1}, we can in principle derive an alternative way to obtain disentangled and contextualized attributions with respect to $\widetilde{y}$. Instead of adding up the respective expert attributions, we subtract them from the original explanation in reverse order. Intuitively, only information relevant to higher-range bins should remain. We test this hypothesis empirically on the \textit{rMNIST} dataset. We flip pixels to zero according to the order of the difference of attributions $\mathcal{E}(y,x)-\sum_{m}{\mathcal{E}(z_m,x)}$. Intuitively, the more evidence associated with lower-range concepts we subtract, the more evidence for higher values should remain, and therefore the flipping curve should decrease more slowly. In the ideal case, the only information with positive attributions is the one relevant for values larger than $\widetilde{y}_i$ and the flipping curve should therefore remain around that value for as long as possible. When testing this empirically, we indeed see such a behaviour (Figure \ref{fig:flip_curves}). Note also, that the plateaus of the different range experts do not cluster around the digit-transitions (0.33 and 0.66). This means, that despite the strong global concept shifts present in the data, the range experts were able to capture more subtle, local effects that guide $f(x)$ in the context of the respective reference values.

\section{Discussion and Conclusions}
\label{sec:conclusions}

In this paper, we have proposed the \textit{XpertAI} framework to achieve contextualized and disentangled attributions when explaining regression models. Inspired by the MoE approach, the framework divides the data into sub-manifolds, each of which corresponds to a certain predicted output range. 
Such a division is achieved by building a collection of \textit{range experts}, which we equip with explainability. It enables for the first time a disentanglement along the output of the prediction strategy and the resolution of specific user-defined queries.

Empirically, we find that our XpertAI framework can distill locally relevant explanations from highly aggregated global standard attributions, as demonstrated by several quantitative experiments based on occlusion tests. Explanations associated with each expert range can be precomputed, so that exact user queries can be answered very quickly as a linear combination of the precomputed explanations.

Our approach can be interpreted within the framework of virtual layers, which has been instrumental in achieving various forms of explanation disentanglement. Furthermore, our approach provides an alternative to the more common approach of extracting reference points or counterfactuals and bypasses some of the challenges , such as their multiplicity and the need to search for them. Also, our approach differs from self-interpretable generalized additive models, by remaining applicable to a broad range of ML models, including deep neural networks.

We have demonstrated that our method can work alongside various explanation techniques, in particular, gradient-based techniques such as Integrated Gradients, or propagation-based techniques such as LRP. While this enables a seamless integration into existing explanation pipelines our approach naturally inherits potential shortcomings of these methods. Furthermore, it is necessary for propagation-based techniques to structurally disentangle the range experts. While we have proposed a surrogate modeling approach for this step, these surrogates need to be carefully trained and regularized to maintain the original model's prediction output as well as its prediction strategy. Also, retraining implies additional computational cost. Hybrid approaches, with the top layers handled by perturbation-based techniques and the lower layers with propagation, may eliminate the need for structural disentanglement while at the same time retaining high accuracy and computational efficiency. Enhanced approaches, inspired by model distillation or formally equivalent neural networks, could also be considered.

Overall, our work has highlighted the need to precisely formulate ``what to explain'' (the explanandum) and proposed a practical and flexible solution in the context of regression. The MoE idea our method builds upon,  however, is more general, and our framework could be extended in the future to other decomposition of the predicted output, e.g.\ for structured output tasks such as time series prediction. Additional future work could furthermore focus on automating the optimal number of experts in a data-driven way. While we have shown that for sufficiently populated ranges of the output adding more experts improves contextualization, there certainly are limitations arising from data availability and computational constraints. Lastly, the application and evaluation to more complex models, such as regression foundation models \cite{hollmann2025accurate}, should be considered in the future.

\subsubsection{\ackname} This work was partly funded by the German Ministry for Education and Research [01IS24087C, 01IS14013A-E, 01GQ1115, 01GQ0850, 01IS18056A, 01IS18025A, and 01IS18037A], the German Research Foundation as Math+: Berlin Mathematics Research Center [EXC2046/1, project-ID: 390685689], the Investitionsbank Berlin [10174498 ProFIT program], and the European Union’s Horizon 2020 Research and Innovation program under grant [965221]. Furthermore, Klaus-Robert Müller was partly supported by the Institute of Information and Communications Technology Planning and Evaluation grants funded by the Korean Government [2019-0-00079].
Our gratitude extends to Jonas Lederer, Pattarawat Chormai and Stefan Blücher for their invaluable comments and feedback that have contributed
to enhancing the quality of the manuscript.

%
%
%
\newpage

\bibliographystyle{splncs04}
\bibliography{bibliography}

\newpage

\appendix

\section{Proof of Propositions 1 and 2}
\label{app:proof_proposition}
Proposition \ref{proposition:conservation} stating the conservation property of the proposed query explanation can be demonstrated through the chain of equations:
\begin{align}
\textstyle
\sum_i \mathcal{E}(q,x)_i
&= \textstyle \sum_i \sum_m w_m \mathcal{E}(z_m,x)_i \label{eq:proof-step1}\\
&= \textstyle \sum_m w_m \sum_i \mathcal{E}(z_m,x)_i\label{eq:proof-step2}\\
&= \textstyle \sum_m w_m z_m\label{eq:proof-step3}\\
&= q \label{eq:proof-step4}
\end{align}
where in \eqref{eq:proof-step1}, we have injected the expression of the explanation in \eqref{eq:explanation-sum}. From \eqref{eq:proof-step1} to \eqref{eq:proof-step2} we have permuted the sums. From \eqref{eq:proof-step2} to \eqref{eq:proof-step3}, we have used the conservation property of the explanation of $z_m$. From \eqref{eq:proof-step3} to \eqref{eq:proof-step4} we have identified the weighted sum as being the query. Likewise, for Proposition \ref{proposition:irrelevance}, if some feature $i$ satisfies $\forall_m: \mathcal{E}(z_m,x)_i = 0$, then
\begin{align}
\mathcal{E}(q,x)_i
&= \textstyle \sum_m w_m \mathcal{E}(z_m,x)_i\\
&= \textstyle \sum_m w_m \cdot 0\\
&= \textstyle 0
\end{align}

\section{How to train and select good range experts?}
\label{app:best_practice}
In practice, we need to select appropriate \textit{range experts} for the XpertAI approach to enhance contextualization. This process may vary based on the respective XAI attribution method being employed. For occlusion- and gradient-integration-based methods, which do not require additional structural disentanglement (see Section \ref{section:technical}), a simple shift-and-clip strategy is sufficient. For propagation-based methods, however, we need to learn the surrogate $a \mapsto (z_m)_m$ (see Sec. \ref{sec:method}). Here, we want to highlight the need for appropriate regularization to avoid overfitting, which in the case of range experts would result in unfaithful model attributions. Analogously to regular model selection, we aim to choose the least complex range expert, that can sufficiently learn the respective mapping.

In case the latent representation $a$ is already adequately disentangled, it is sufficient to fit a linear range expert (without bias term). We have observed this to work well for some of our low-dimensional benchmark datasets. Otherwise, we need to gradually increase range-expert complexity (adding neurons and reducing L2-regularization) until the mapping is learned sufficiently. Moreover, we have observed that instead of additional layers, (copying and) fine-tuning the top layer(s) on the range-expert targets $z_m$ with small learning rates is a good strategy since it ensures the solution lays in relative proximity to the original model. If a new layer is added, initializing the weights with a projection to the latent principal components conditioned on the respective output range ($PCA(X|z_m)$) was found to speed up training and ensure good results. Furthermore, the Shapley-style data augmentation (cf.\ Section \ref{sec:method}) is another crucial ingredient to prevent our experts from adhering to spurious correlations (that our original models did not use). This can be conveniently implemented with the help of a dropout layer on the surrogate input $\boldsymbol{a}$. Lastly, we can enforce the saturation of range experts outside their area of expertise by adding an explicit combination of ReLU functions that clip $s_m$ to the desired range. These measures together ensure faithful and computationally efficient \textit{range experts}.

\section{Details on Evaluation (Sec. \ref{sec:qual_eval} and \ref{sec:quant_eval})}
\label{app:details_case}

\subsection{Details face-age regression example}
For this analysis, we have made use of a dataset containing $\sim$ 20k facial images associated with biological age\footnote{\url{https://www.kaggle.com/frabbisw/facial-age}} (biased toward younger ages). Each image is pre-processed so that all of them have the same size (200x200) and the faces are aligned and centred. We used a VGG-16 \cite{DBLP:journals/corr/SimonyanZ14a} model pre-trained on ImageNet \cite{DBLP:conf/cvpr/DengDSLL009,DBLP:journals/ijcv/RussakovskyDSKS15} as a feature extractor followed by one ReLU layer with 256 neurons, a dropout-layer, and a final linear layer mapping the 256 neurons to a real-valued age prediction. In all cases we used \mbox{LRP-$\alpha_1\beta_0$} rule \cite{bach-plos15,montavon2018methods} in the convolutional layers and \mbox{LRP-$\epsilon$} rule \cite{bach-plos15} (where biases are ignored) for the fully connected layers.

\subsection{Details quantitative evaluation}
Here, we describe the details of our quanitative experiments. For the rMNIST experiments, we utilized a vanilla CNNs with two convolutional, ReLU and pooling layers, followed by three fully connected layers. The convolutional blocks were kept frozen, and only the fully connected layers were re-trained as experts, starting from their original model weights. The other problems (Wind, Wine, California and Diabetes) are based on tabular data. Here, we utilize a 4 layer-MLP with 20 neurons in each hidden layer. The last two were re-trained for each expert. Moreover, we utilized the PCA initialization trick, described above (Appendix B). $w_m$ was selected to be 1 for all expert ranges between reference value and sample output, and 0 otherwise. More specific information on the implementation can be found in the published code repository.\footnote{https://github.com/sltzgs/XpertAI}

\begin{figure}[h!]
\centering
\includegraphics[width=0.5\linewidth]{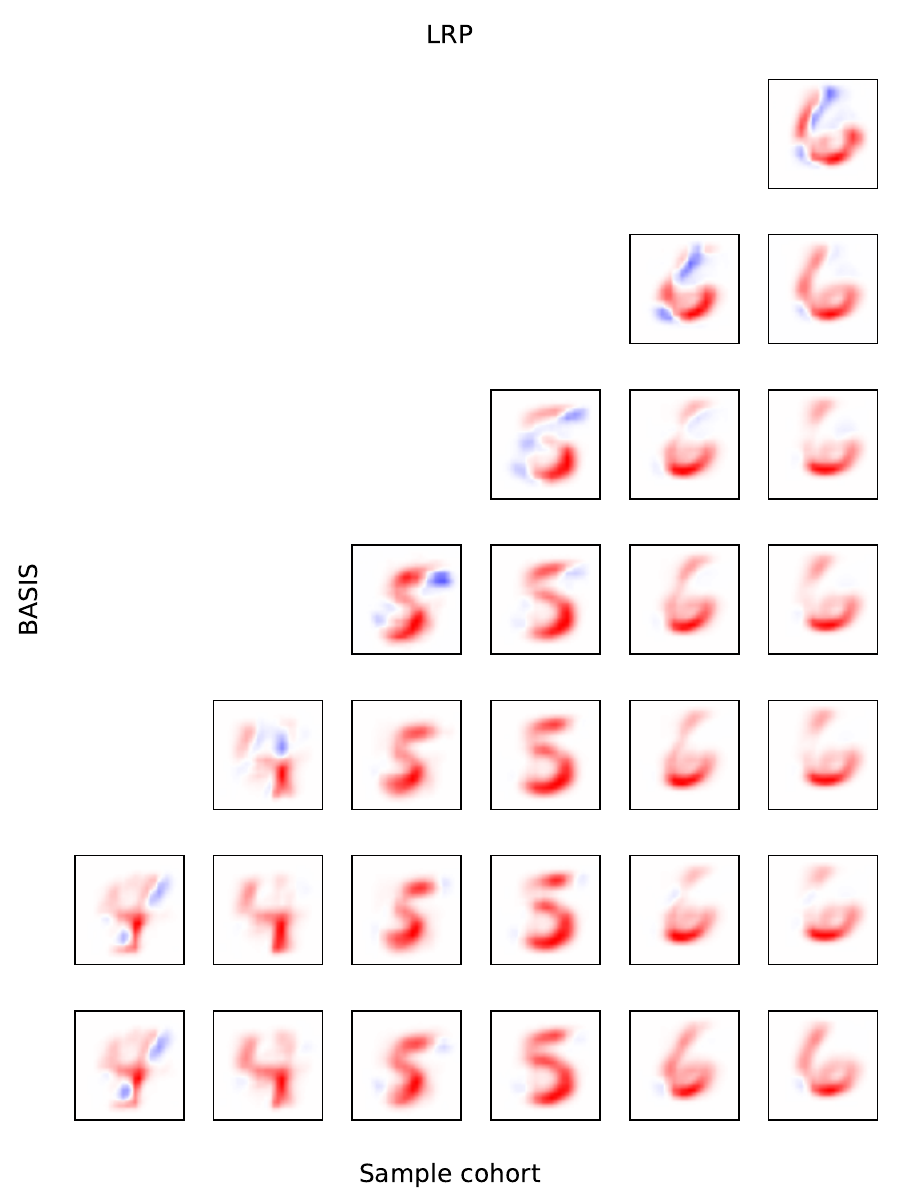}
    \caption{Mean attributions over different cohorts of samples (columns) and basis functions (rows). Equivalent plot to Fig. \ref{fig:mean_att_mnist_three} but for six range expert basis functions.}
\end{figure}

\begin{figure}[h!]
    \centering
    \includegraphics[width=.75\linewidth]{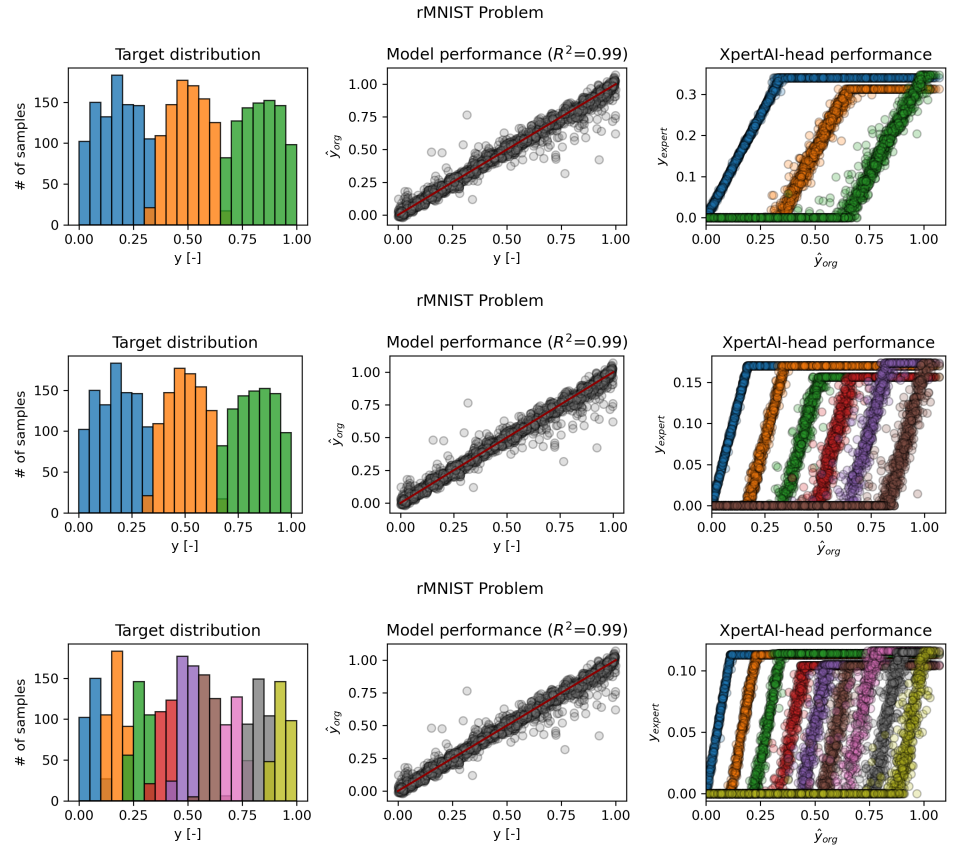}
    \caption{Overview of model performance on the rMNIST problem for 3, 6 and 9 range experts (top to bottom).}
    \label{fig:overview_rmnist}
\end{figure}


\begin{figure}[h!]
    \centering
    \includegraphics[width=.75\linewidth]{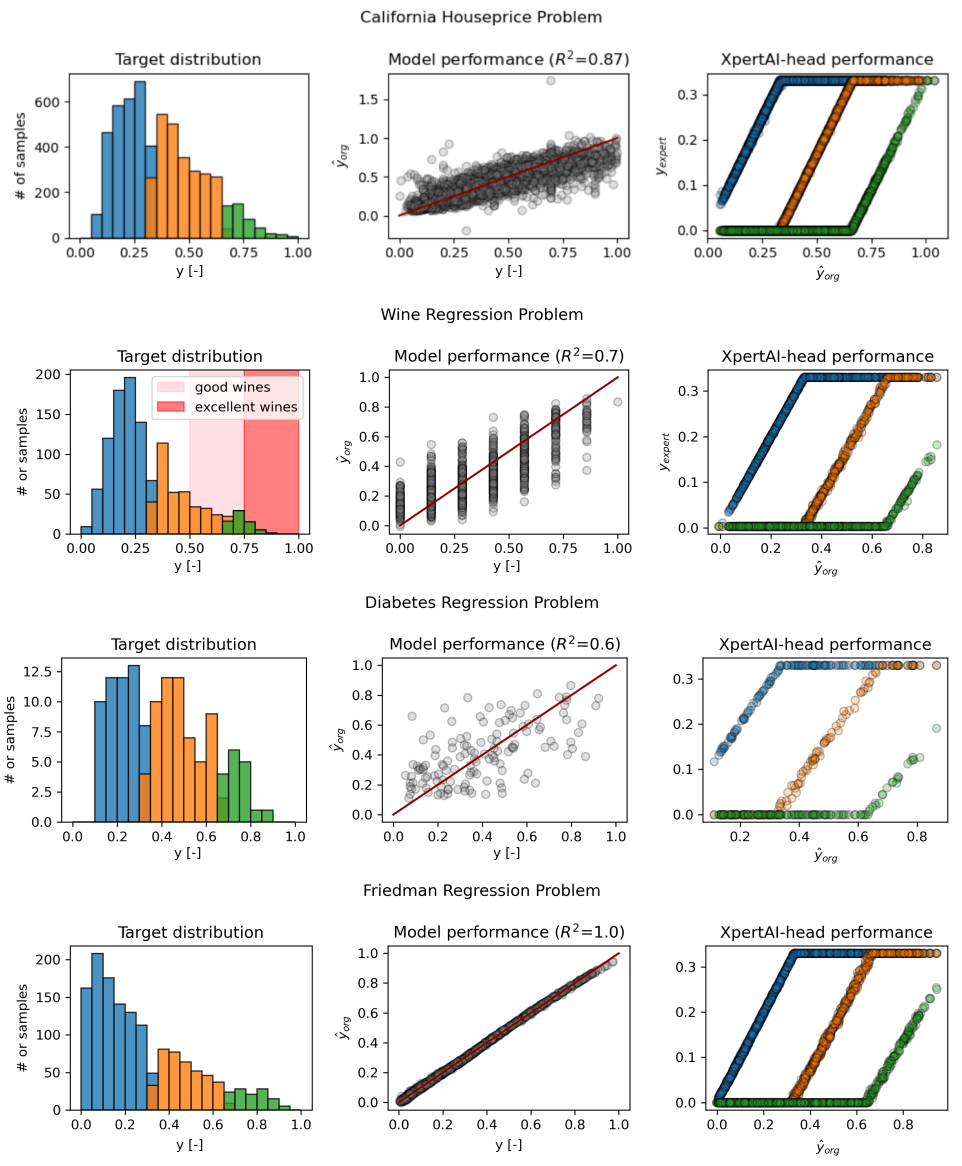}
    \caption{Overview model performance for the regression benchmarks (Sec. \ref{sec:quant_eval}).}
    \label{fig:overview_benchmarks}
\end{figure}

\subsection{Augmenting wind turbine SCADA data with yaw-misalignment}
\label{app:details_cs_wind}
We randomly add yaw misalignment of up to $15 ^\circ$ to our data SCADA set, and adjust the respective targets (turbine output) with a yaw misalignment factor $c_{ymis,i} = cos^3(\Delta_{yaw})$, if $v_{w,i}<v_{w,rated}$. This approximation can be easily derived from static flow equations and geometric considerations, for more details on how yaw misalignment affects turbine output see \cite{howland2020influence}. After training and evaluation of the model on the augmented data, we can compare the magnitude of attributions to the ground truth.

\begin{figure}[h!]
    \centering
    \includegraphics[width=.75\linewidth]{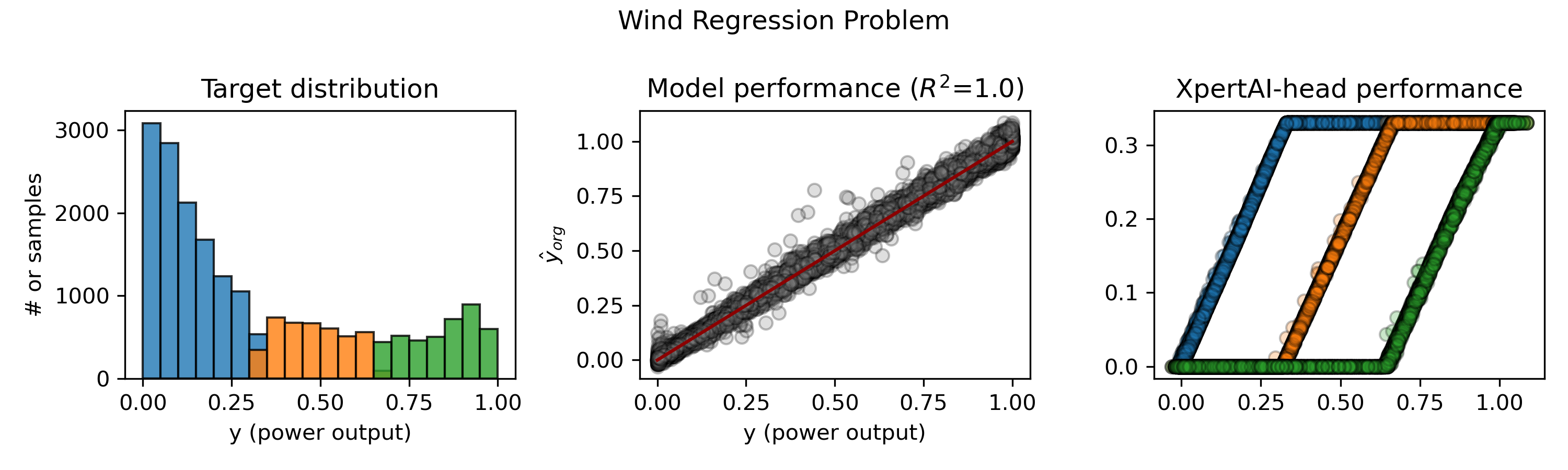}
    \caption{Overview model performance wind turbine example (Sec. \ref{sec:wind_example})}
    \label{fig:overview_benchmarks}
\end{figure}

\end{document}